\PassOptionsToPackage{dvipsnames}{xcolor}

\documentclass[pmlr]{jmlr}

\RequirePackage{graphicx}
 \usepackage{booktabs}
\usepackage{longtable}%

\makeatletter
\def\set@curr@file#1{\def\@curr@file{#1}} %
\makeatother
\usepackage[load-configurations=version-1]{siunitx} %

\theorembodyfont{\upshape}
\theoremheaderfont{\scshape}
\theorempostheader{:}
\theoremsep{\newline}

\jmlrvolume{298}
\jmlryear{2025}
\jmlrworkshop{Machine Learning for Healthcare}

\usepackage[dvipsnames]{xcolor}
\definecolor{myblue}{RGB}{31, 119, 180}
\definecolor{myorange}{RGB}{255, 127, 14}
\definecolor{mygreen}{RGB}{44, 160, 44}
\definecolor{myred}{RGB}{214, 39, 40}
\definecolor{myyellow}{RGB}{230,194,0}

\usepackage[noend]{algpseudocode}
\usepackage{algorithm}

\usepackage{cite}
\usepackage{comment}
\usepackage{prettyref}
\usepackage{amsfonts}
\usepackage{makecell}
\usepackage{adjustbox}
\usepackage{multicol}

\newcommand{\equal}[1]{{\hypersetup{linkcolor=black}\thanks{#1}}}

\usepackage{amsmath}
\usepackage{amssymb}
\usepackage{mathtools}

\usepackage[colorinlistoftodos]{todonotes}

\usepackage{xspace}
\usepackage{enumitem}

\newcommand\Nodes{\mathcal{V}}

\newcommand{\algname}{\text{Bi-Hierarchical Fusion}\xspace}

\newcommand{\fwname}{\textsc{Bi-Hierarchical Fusion}\xspace} %

\newif\iffinal
\finaltrue

\iffinal
    \newcommand{\fix}[1]{#1}
    \newcommand{\pref}[1]{}
    \newcommand{\XL}[1]{}
    \newcommand{\CT}[1]{}
    \newcommand{\SJ}[1]{}

\else
    \newcommand{\fix}[1]{{\color{red} #1}}
    \newcommand{\XL}[1]{\todo[fancyline,color=Maroon!40]{XL: #1}\xspace}
    \newcommand{\SJ}[1]{\todo[fancyline,color=Fuchsia!40]{SJ: #1}\xspace}
    
    \newcommand{\CT}[1]{\todo[fancyline,color=Fuchsia!40]{CT: #1}\xspace}

    \newcommand{\pref}[1]{{\color{blue}(\ref{#1})}}
\fi

\newcommand{\pLMs}{\ensuremath{{\text{pLMs}}}}

\newcommand{\tabref}[1]{Table~\ref{#1}}
\newcommand{\figref}[1]{Fig.~\ref{#1}}

\newcommand{\secref}[1]{\S\ref{#1}}

\newcommand{\paren} [1] {\ensuremath{ \left( {#1} \right) }}

\newcommand{\bracket}[1]{\left[#1\right]}

\newcommand{\curlybracket}[1]{\ensuremath{\left\{#1\right\}}}

\title[\algname]{Bidirectional Hierarchical Protein Multi-Modal Representation Learning}

\author{\Name{Xuefeng Liu}\equal{These authors contributed equally.\\ X. Liu and R. Stevens are also affiliated with Argonne National Laboratory.} 
       \Email{xuefeng@uchicago.edu}\\ 
       \addr Department of Computer Science\\
       University of Chicago\\
       Chicago, IL, U.S. 
       \AND
       \Name{Songhao Jiang}\footnotemark[1]
       \Email{shjiang@uchicago.edu}\\ 
       \addr Department of Computer Science\\
       University of Chicago\\
       Chicago, IL, U.S. 
        \AND
       \Name{Chihchan Tien}
       \Email{cctien@uchicago.edu }\\ 
       \addr Department of Computer Science\\
       University of Chicago\\
       Chicago, IL, U.S. 
        \AND
        \Name{Jinbo Xu}
       \Email{jinboxu@gmail.com}\\ 
       \addr Toyota Technological Institute at Chicago\\
       Chicago, IL, U.S.  
        \AND
        \Name{Rick Stevens}
       \Email{stevens@cs.uchicago.edu}\\ 
       \addr Department of Computer Science\\
       University of Chicago\\
       Chicago, IL, U.S. 
       } 

\begin{document}

\maketitle

\begin{abstract}

Protein representation learning is critical for numerous biological tasks.
Recently, large transformer-based protein language models ($\pLMs$) pretrained on large scale protein sequences have demonstrated significant success in sequence-based tasks. However, $\pLMs$ lack structural context, and adapting them to structure-dependent tasks like binding affinity prediction remains a challenge.
Conversely, graph neural networks (GNNs) designed to leverage 3D structural information have shown promising generalization in protein-related prediction tasks, but their effectiveness is often constrained by the scarcity of labeled structural data. Recognizing that sequence and structural representations are complementary perspectives of the same protein entity, we propose a \fix{multimodal} \fix{bidirectional hierarchical} fusion framework to effectively merge these modalities.
Our framework employs attention and gating mechanisms to enable effective interaction between $\pLMs$-generated sequential representations and GNN-extracted structural features, improving information exchange and enhancement across layers of the neural network. 
This \fix{bidirectional and hierarchical (Bi-Hierarchical)} fusion approach leverages the strengths of both modalities to capture richer and more comprehensive protein representations.
\fix{Based on the framework, we further introduce local Bi-Hierarchical Fusion with gating and global Bi-Hierarchical Fusion with multihead self-attention approaches.
}
Through extensive experiments on a diverse set of protein-related tasks, our method demonstrates consistent improvements over strong baselines and existing fusion techniques in a variety of protein representation learning benchmarks, including 
react (enzyme/EC classification), 
model quality assessment (MQA), 
protein-ligand binding affinity prediction (LBA), 
protein-protein binding site prediction (PPBS), and
\fix{B cell epitopes prediction (BCEs)}. 
Our method establishes a new state-of-the-art for multimodal protein representation learning, emphasizing the efficacy of 
\fix{\fwname}
in bridging sequence and structural modalities.

\end{abstract}

\section{Introduction}
\label{sec:introduction}

\begin{figure*}[t]
    \centering
    \includegraphics[width=1\linewidth]{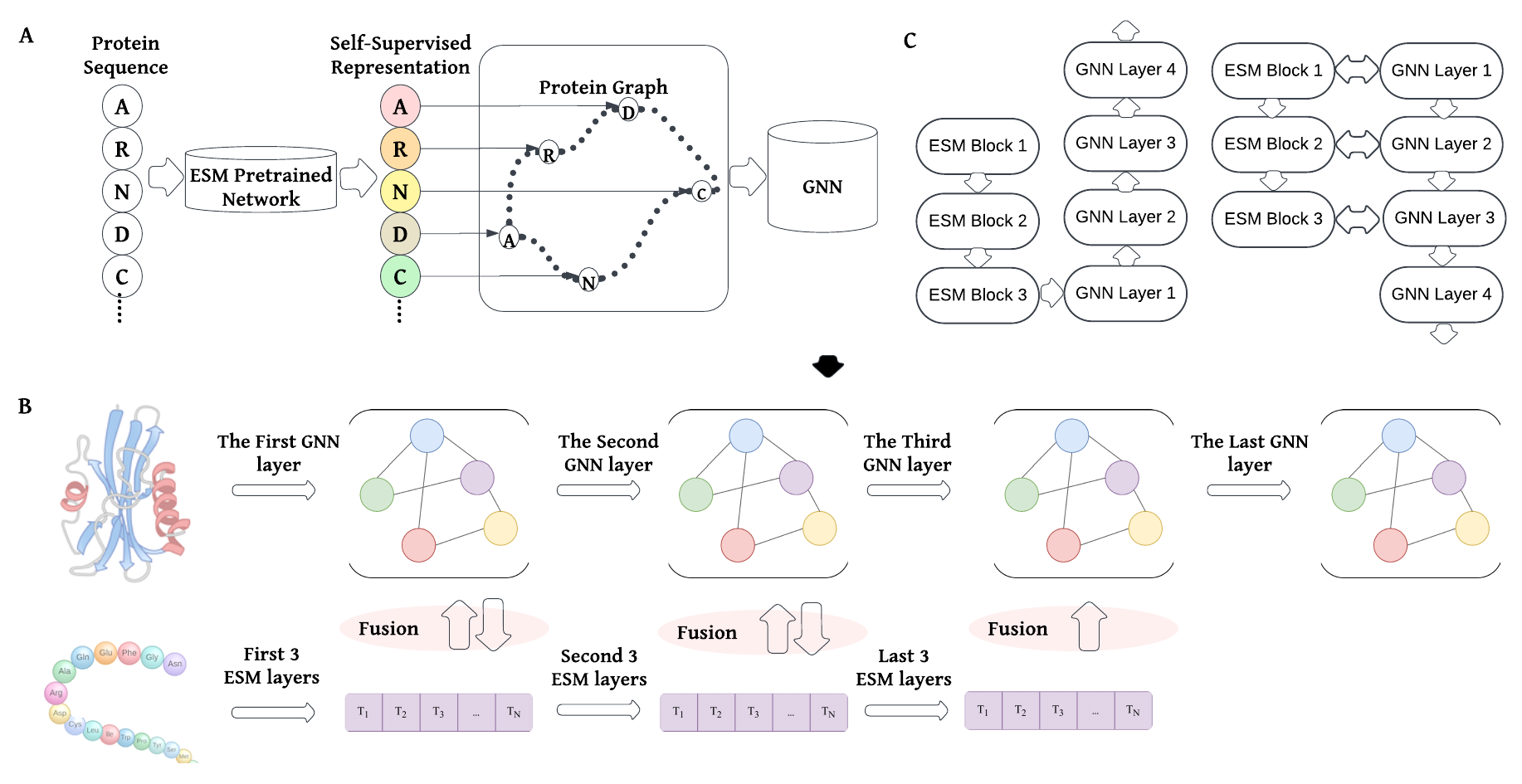}
    \caption{Overview of serial fusion (A), comparative overview of serial fusion and \fix{bi-hierarchical} fusion (B), and our \algname (C). 
    \textbf{(A) Serial Fusion Framework.} The protein sequence is processed through the pre-trained protein language model, ESM, to generate per-residue representations. These representations are then employed as node features within 3D protein graphs for subsequent analysis by the baseline GNN, ProNet.
    \textbf{(B) \fix{\fwname} Framework.} The proposed structure is a two-branch network, characterized by intricate interactions among its branches. Specifically, the sequence-branch (below) leverages ESM, and the graph-branch (above) employs the selected baseline GNN, ProNet. This schema applies to both the local \algname with gating and the global \algname with multihead attention. \textbf{(C) Comparison of Serial Fusion and \fix{Bi-Hierarchical} Fusion.} }
    \label{fig:main1}
\end{figure*}

Proteins are essential building blocks of life. While proteins can be represented as one-dimensional sequential data, their complex three-dimensional structures and dynamic nature underscore their vast functional diversity. 
A thorough understanding of protein 3D structures is critical for unraveling disease mechanisms and advancing drug discovery. 
Consequently, extensive research has been conducted on protein 3D structure representation learning, demonstrating its effectiveness across diverse protein analysis tasks~\citep{baldassarre2021graphqa, wang2023pronet, yang2023geometric}.
With advancements in deep learning, 3D geometric graph neural networks (GGNNs) have been developed to model protein structural information, yielding significant improvements in prediction tasks involving proteins \citep{fan2022continuous, zhang2022gearnet, wang2023pronet, wu2023integration}. However, the limited availability of labeled data constrains the power of GGNNs. 
In addition, existing GGNNs are proven to be unaware of the positional order within the protein sequence~\citep{wu2023integration}.

\begin{figure*}[t]
    \centering
    \includegraphics[width=1\linewidth]{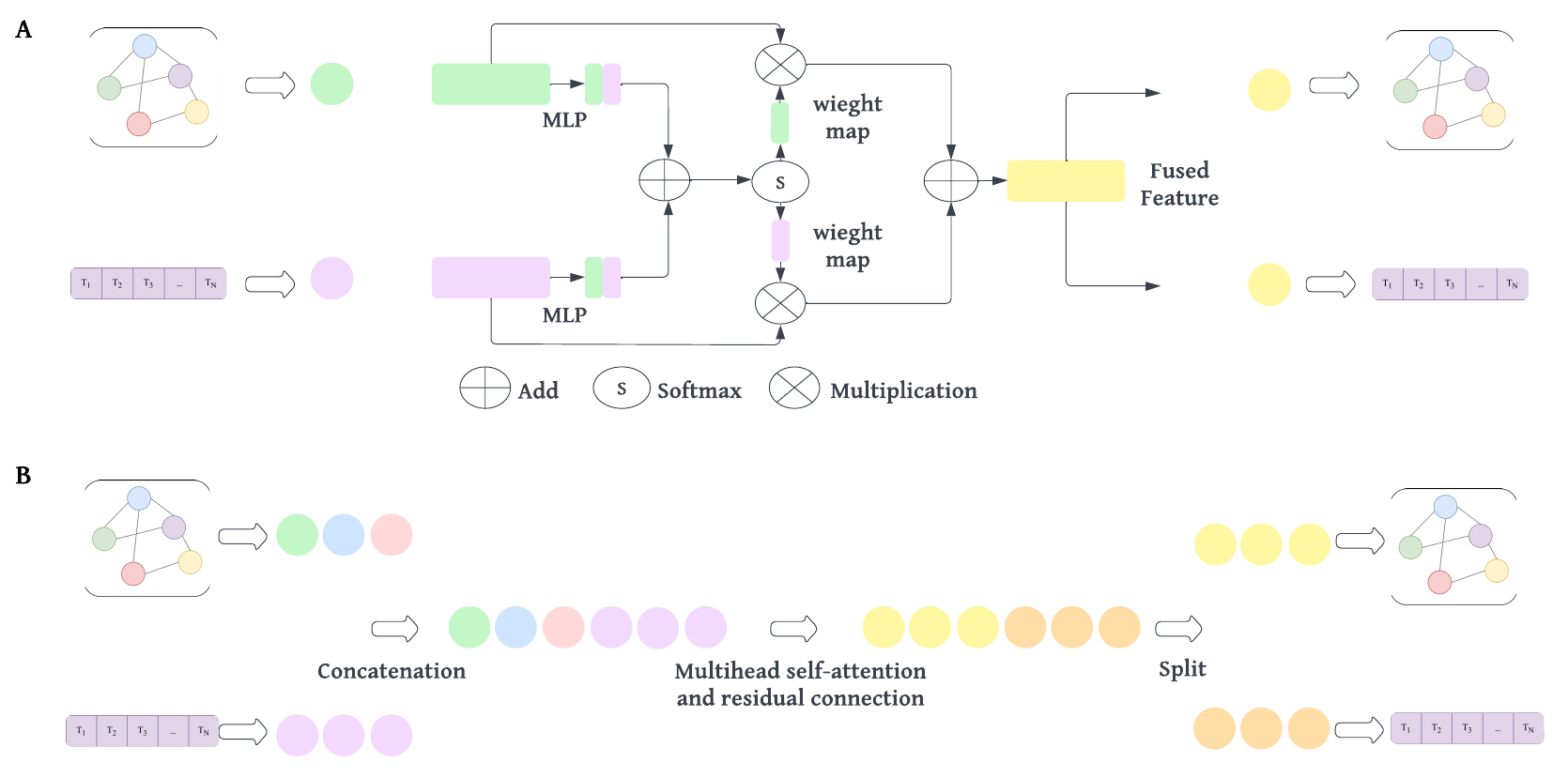}
    \caption{\textbf{Overview of local \algname with gating and global \algname with multi-head attention.} 
    \textbf{(A) Local \algname with gating.} In the left block: Given an amino acid, we need to find the corresponding character in sequence and graph node. In the central block: When presented with features of an amino acid from varied representations, we merge them adaptively with gating mechanism. In the right block: Once the features are fused, the next step is to map them back to their respective representations.
    \textbf{(B) Global \algname with multihead self-attention.}  For each protein, we concatenate representations from GNN (above) and $\pLMs$ (below) along the dimension of nodes/tokens, and perform multi-head self-attention over the newly concatenated sequence. The resulting representations are then split and put back to the respective graphical and sequential structure.
    }
    \label{fig:main2}
\end{figure*}

On the other hand, protein folding models \citep{jumper2021highly, lin2023esm}, which predict 3D structures from protein sequences, highlight the rich information embedded in one-dimensional sequential data. 
Inspired by the success of large pretrained language models (LLMs) in natural language processing \citep{radford2019language, raffel2020exploring}, researchers have adapted LLMs for protein representation learning using protein sequences. These protein language models ($\pLMs$) treat protein sequences as a language, with individual amino acids as tokens. Prominent advancements in this area include UniRep \citep{alley2019unified}, ProtTrans \citep{elnaggar2021prottrans}, and ESM \citep{lin2023esm}.
Although $\pLMs$ benefit from the versatile Transformer architecture and the relatively greater availability of unlabeled sequence data, a model based on sequence prediction alone lacks the structural information and hierarchical representation of proteins, which 
restricts the utilization of labeled structural data and may lack the inductive bias to 
represent proteins consistent with physical and chemical constraints. Thus, adapting $\pLMs$ to tasks involving structural input, such as protein structure and protein-protein interaction prediction remains challenging. 

Therefore, integrating diverse modalities of data representation offers a promising avenue to enrich protein analysis.  The serial fusion framework proposed in previous work \citep{wu2023integration, zhang2023gearnet_esm, zhang2023systematic} is one way to combine representations of $\pLMs$ and GNNs for supervised learning applications, as depicted in \figref{fig:main1}(A). Initially, protein sequences are processed through a $\pLMs$ to generate detailed per-residue representations. These representations are then utilized as node features in 3D protein graphs, which are further analyzed using a GNN. This integration ensures that the information captured by $\pLMs$ enrich the structural analysis performed by GNN, potentially leading to more accurate and insightful predictions.

A notable example of serial fusion is ESM-GearNet model~\citep{zhang2023gearnet_esm}. This model incorporates the output of the ESM into GearNet~\citep{zhang2022gearnet}, by substituting GearNet's node features with those derived from ESM. The resultant representation benefits from the deep evolutionary insights encoded by $\pLMs$, demonstrating the potential of combining $\pLMs$ with GNNs for advanced protein representation. However, the reliance of ESM-GearNet on self-supervised learning for pre-training poses questions about its adaptability and efficacy in supervised learning contexts. \XL{revisit}
Furthermore, there are two inadequacies of the serial fusion method as shown in \figref{fig:main1}(A).
One is that $\pLMs$ do not receive the structural information from GNN, Thus the interaction between two branches is only unidirectional.  
The second is that the exchange of information between two branches happens only once, which may limit how much the system can fully benefit from different but complementary views of the same object \fix{at different hierarchical layer}. 

\emph{\algname Architecture.}
To address these drawbacks of serial fusion, in this work, we propose the \fix{\algname architecture}, which integrates protein sequence and graph representations bidirectionally. Moreover, considering various layers of $\pLMs$ and GNN could accumulate different levels of information, we combine these representations in a \fix{hierarchical manner}.
Our proposed framework, as depicted in \figref{fig:main1}(B), integrates sequence and graph representations, leveraging multiple and mutual information interactions in a bidirectional manner. This framework aims to capture a complete knowledge of the protein from various perspectives, with each representation providing unique insights to enhance the model's predictive accuracy.  
Within the architecture, we introduce
\fix{local \algname with gating} and \fix{global \algname with multihead self-attention approaches.}

\emph{Local \algname with gating.} 
To effectively combine features from these representations, we utilize a gating mechanism, as shown in Figure \ref{fig:main2}(A). This method dynamically adjusts the weight of each modality's contribution, improving the integration process and clarifying the influence of each modality on the final prediction. The gated fusion layer modifies the traditional addition-based fusion approach by controlling the information flow through gates, drawing inspiration from RPVNet~\citep{xu2021rpvnet}. This technique not only enables seamless integration of sequence and graph data but also enhances the model’s interpretability by highlighting the contributions of different modalities on the final outcomes. See \secref{sec:local} for more details.

\algname techniques enable nodes to \fix{bidirectionally and hierarchically} integrate perspectives from both GNN and $\pLMs$. However, the local \algname as described only facilitates information exchange between nodes that correspond to the same amino acid within the protein structure. We propose that the model could also gain from allowing a node to receive information from other nodes in a different branch, not just from the corresponding node. 

A clear instance of why an amino acid might need information from other amino acids or a sequence thereof arises in the formation and stabilization of protein structures via hydrogen bonding. An amino acid must be aware of others several residues away to maintain structural integrity and functionality. This structural awareness is crucial because it influences the protein's ability to engage with other molecules and execute its biological roles. This scenario illustrates the importance of both local and remote amino acid interactions in protein chemistry.
Moreover, each node might benefit from various combinations of representations that together enhance subsequent representations.

\emph{Global \algname with Multi-head Attention.}
To address the limitation and enhance communication across branches, we introduce Global \algname with Multi-head Attention, as illustrated in Figure \ref{fig:main2}(B). This approach utilizes multi-head self-attention across nodes and tokens, enabling each node and token in both branches to potentially engage with any combination of nodes or tokens from either branch. This method overcomes the limitations of Local \algname, which restricts information exchange to identical nodes or amino acids. Unlike locally gating \algname, which produces a uniform representation, global fusion allows for branch-specific output representations. This variability arises because different branches may require distinct information; for example, the pretrained LLMs branch might need structural or hierarchical data from the GNN branch, which the GNN branch does not require. Finally, we integrate a residual connection following the multi-head attention module. See \secref{sec:global} for more details.

\subsection*{Generalizable Insights about Machine Learning in the Context of Healthcare}
\fix{Enhancing protein representation has direct impact on drug discovery, structural biology, and immunotherapy, and plays a crucial role in advancing AI applications in clinical research and practice, including drug and vaccine development.}
\fix{In this work, we introduce the \algname framework for protein representation learning. Building on this framework, we develop two variants: local \algname with gating mechanisms and global \algname with multi-head attention.}
We evaluate our methodologies across a diverse array of benchmarks, including model quality assessment, protein-ligand binding affinity, reaction prediction, protein-protein binding site prediction, and \fix{B cell epitopes (BCE) prediction}. Our comprehensive experiments demonstrate that our \algname approach, which includes token-wise and global information exchange, surpasses the previous state-of-the-art \fix{in protein representation learning}, Pronet, and serial fusion across various tasks that require both structural and sequential knowledge. This underscores that our \algname technique facilitates more effective knowledge exchange between different branches of protein representation compared to the serial fusion approach. Our findings contribute to a better understanding of how to effectively utilize burgeoning geometric deep learning, well-established protein language models, and enhance the integration of various protein modalities.

Our contribution are threefold:
\begin{itemize}
  \setlength\itemsep{1pt}
    \item We design the innovative fusion architecture, \algname, which \fix{bidirectionally and hierarchically} merges the representations from the large protein language models ($\pLMs$) and the graph neural networks (GNNs) to facilitate the learning of multi-modal protein representations.

    \item 
    \fix{Building on the \algname architecture, we further introduce two fusion methods: local bi-hierarchical fusion with gating, which facilitates information exchange between nodes of the same amino acid within the protein structure, and global bi-hierarchical fusion with multihead attention, which allows for information exchange between different nodes in distinct branches. Both methods surpass the current state-of-the-art, serial fusion, in performance.}
    
    \item Finally, we conduct experiments on the benchmark datasets on various tasks, covering from single protein representation, protein-molecule representation, and protein-protein representation. We demonstrate superior performance of our approach on protein representation learning, compared with serial fusion and other STOA methods.
\end{itemize}

\section{Method}
\subsection{Background}

\textbf{Notations.} 
To model protein representations with 3D structures, we represent a protein as a 3D graph $G = (\Nodes, \mathcal{E}, \mathcal{P})$. 
Here, $\mathcal{V}=\curlybracket{{\mathbf{v}_i}}_{i=1,...,n}$ represents the set of node features, 
with each $\mathbf{v}_i\in \mathbb{R}^{d_v}$ indicating the feature vector for node $i$.
$\mathcal{E}=\curlybracket{e_{ij}}_{i,j=1,\cdots,n}$ comprises the set of edge features, where $e_{ij}\in \mathbb{R}$ corresponds to the feature vector for edge $ij$. 
$\mathcal{P}=\curlybracket{P_i}_{i=1,\cdots,n}$ denotes the set of position matrices, with each $P_i\in \mathbb{R}^{k_i\times 3}$ representing the position matrix for node $i$.
The value of $k_i$ varies across different applications.
For instance, in the context of molecules where each atom is considered a node, $k_i$ is 1 for each node $i$.
Conversely, in proteins where each amino acid is treated as a node, 
$k_i$ corresponds to the number of atoms in amino acid $i$.

In this graph, each amino acid is a node, and edges are established between nodes if the distance between them is less than a certain cutoff radius. 
Each node $i$ has a feature $\mathbf{v}i$, which is a one-hot encoding of its amino acid type. 
Each edge $ij$ has a feature $\mathbf{e}_{ij}$, representing the embedding of the sequential distance $j-i$, consistent with prior research. 
Additionally, the position matrix $P_i$ for a node $i$ contains the coordinates for all atoms of the amino acid when available, arranged in a predefined atom order. For instance, in the amino acid alanine, the sequence of atoms in the position matrix is $N, C_{\alpha}, C, O$, and $C_{\beta}$.

\textbf{Complete Geometric Representations.} 
As described in~\citep{wang2022comenet}, a geometric transformation  $\mathcal{F}\paren{\cdot}$  is considered complete if it holds that for any two 3D graphs $G^1=\paren{\mathcal{V},\mathcal{E},\mathcal{P}^1}$ and $G^2=\paren{\mathcal{V},\mathcal{E},\mathcal{P}^2}$, the condition
 $\mathcal{F}\paren{G^1}=F\paren{G^2} \Leftrightarrow  \exists R \in SE\paren{3}$ implies there exists a transformation $ R \in SE\paren{3}$ such that for every $i$ from 1 to $n$, $P_i^1 = R\paren{P^2_i}$. The group $SE\paren{3}$ represents the Sepecial Euclidean group that accounts for all possible rotations and translations in three dimensions. 
A complete geometric representation inherently possesses rotation and translation invariance, reflecting the natural properties of proteins and offering a robust framework for analyzing protein structures.

\textbf{Complete Message Passing Scheme.} 
Incorporating complete geometric representations into the widely used message passing framework enables us to formulate a full message passing scheme expressed as 
$\mathbf{v}_i^{l+1}=\text{UPDATE}\paren{\mathbf{v}^l_i, \sum_{j\in N_i}\text{MESSAGE}\paren{\mathbf{v}^l_j,\mathbf{e}_{ji},\mathcal{F}\paren{G}}}$, 
where $\mathcal{N}_i$ indicates the set of neighbors for node 
$i$. The UPDATE and MESSAGE functions are achieved using neural networks or mathematical operations.
With this representation and a complete message passing scheme, a comprehensive representation for a whole $3D$ protein graph is achieved.

\subsection{Leveraging Protein Large Language Models and Graph Neural Networks 
}

In this study, we present a novel framework for integration of pre-trained protein large language model and graph neural network.
Our model introduces innovative fusion methods for integrating protein sequences and structures to refine protein representation.
Specifically, we investigate the potential of transformer-based language models, pre-trained on protein sequences, to augment the SOTA performance of GNNs in this domain.

For our framework, we select ESM \citep{lin2023esm} as the transformer-based protein language model, ProNet \citep{wang2023pronet} for protein GNN.
\XL{ , and HoloProt \citep{somnath2021holoprot} as the molecule GNN.}
ESM is chosen for its proven excellence in tasks such as protein structure prediction with ESMFold \citep{lin2023esm} and protein function prediction in Gearnet-ESM \citep{zhang2023gearnet_esm}. Its capability to encode complex biological data into meaningful representations positions it as an ideal candidate for enhancing protein representation learning and, consequently, benefitting a variety of protein representation learning tasks, such as binding affinity predictions.
ProNet stands out for its geometric representation capabilities at the amino acid level, employing a comprehensive message passing scheme to achieve a full 3D protein graph representation. This aspect is critical for our framework, as ProNet's geometric transformation $F(\cdot)$ ensures SE(3) invariance, essential for maintaining the accuracy of protein representations despite rotations and translations \citep{defresne2021protein}. This invariance is crucial for accurately comparing protein structures by eliminating discrepancies caused by orientation or positional differences.

\textbf{Amino Acid Level Representation.}
Specifically, ProNet designs geometric representation at the amino acid level, $\mathcal{F}\paren{G}_{base}$, as 
$\{(d_{ij}, \theta_{ij}, \phi_{ij}, \tau_{ij})\}_{i=1,\cdots,n, j\in N},$
\fix{where we only consider $C_{\alpha}$ coordinate of each amino acid.}
{In this context, $(d_{ij}, \theta_{ij}, \phi_{ij})$ denotes the spherical coordinates of node $j$ relative to the local coordinate system of node $i$. These coordinates determine the relative position of node $j$, where $d$, $\theta$, and $\phi$ represent the radial distance, polar angle, and azimuthal angle, respectively. Additionally, $\tau_{ij}$ captures the rotation angle of the edge $ji$, accounting for the remaining degree of freedom. Using this representation along with a complete message passing scheme, a detailed and comprehensive representation of an entire 3D protein graph is achieved.}

\textbf{Backbone level Representation.} 
Based on the proposed amino acid level representation, the complete geometric representation at backbone level is $\mathcal{F}\paren{G}_{bb}=\mathcal{F}\paren{G}_{base} \cup \curlybracket{\paren{\tau_{ji}^1,\tau_{ji}^2,\tau_{ji}^3}}_{i=1,\cdots,n, j\in N_i}$, where 
$\tau_{ij}^1,\tau_{ij}^2,\tau_{ij}^3$ are three Euler angles between two backbone coordinate systems.

\textbf{All-Atom Level Representations.} 
An amino acid consists of backbone atoms and side chain atoms. Therefore, building on backbone level representation, we further incorporate side chain information, leading to the all-atom level representation. Based on the backbone level representation, the geometric representation at ALL-Atom level is $\mathcal{F}\paren{G}_{all}=\mathcal{F}\paren{G}_{bb} \cup \curlybracket{\paren{X_i^1,X_i^2,X_i^3,X_i^4}}_{i=1,\cdots,n}$. where $X_i^1,X_i^2,X_i^3,X_i^4$ are first four torsion angles for each amino acid, and the fifth side chain torsion angle is
close to 0.

Here, SE(3) encompasses all possible rotations and translations in a 3D space, introduced to maintain the 3D conformation of a graph despite any rotations and translations, thereby preserving the inherent structure of the graph.
In line with the settings used in ProNet, HoloProt is employed as the ligand network for a fair comparison. Our primary goal is to demonstrate that the integration of transformer models with advanced fusion methods can significantly enhance protein representation learning, thereby improving the accuracy of binding affinity predictions.

Below we present two frameworks, serial fusion and our novel \algname.
These frameworks aim to harness the complementary strengths of each representation type, enhancing the overall predictive power while mitigating their individual limitations.

\textbf{Serial Fusion.}  Serial fusion uses sequence representations as protein residue features in graph neural networks.
Instead of using residue type embeddings to initialize the input node features of the structure encoder, serial fusion leverages the outputs of a protein language model, expressed as 
$\mathbf{u}^{\paren{0}}=\mathbf{h}^{\paren{L}}$. The final protein representations are then obtained from the structure encoder’s output, 
$\mathbf{z}=\mathbf{u}^{L}$.  This method enhances residue type representations by incorporating sequential context, resulting in more expressive features.

\subsection{\algname: Bidirectionally and \fix{Hierarchically} Merging Sequence and Graph Representation}

\subsubsection{Local \algname with Gating}\label{sec:local}

Formally, given two feature vectors \(x^b_i \in \mathbb{R}^{C_b}, i 
\in \bracket{\lvert \mathcal{V} \rvert}, b \in \curlybracket{1,2}, \) from two different branches, where \(\lvert \mathcal{V} \rvert\) is the number of nodes/tokens, and \(C_b\) is the number of channels, 
a multi-layer perceptron \(f\) convert features from both branches to the “votes”.
The gating vector for each node \(i\), \(g_i \in \mathbb{R}^2\) is a softmax on the sum of the votes from both channel, and the final representation is the weighted combination of the two branches according to the gate:
\begin{align*}
     g_i & = \text{softmax}(f(x^1_i) + f(x^2_i))  \\
     \tilde{x}_i &= g_i^1 x^1_i + g_i^2 x^2_i \text{ , }
\end{align*}
where \(g_i^1, g_i^2\) are split from the gate \(g_i\), or \((g_i^1, g_i^2) = g_i\).
Note that the same representation \(\tilde{x}_i\) is then used for both branch.
This is different from the global \algname with multi-head attention which we present next.

\subsubsection{Global \algname with Multi-head Attention}\label{sec:global}

Formally, given a sequence of feature vectors from either of the two branches $b \in \{1, 2\}$, $x_i^b, i 
\in  \bracket{\lvert V \rvert} $,
let \(X^b = \{x^b_i\}|_{i\in \bracket{\lvert  \mathcal{V} \rvert}} = [x^b_1 \, x^b_2 \, \dots \, x^b_{\lvert  \mathcal{V} \rvert}] \) be arrays of representations from one branch, then let \(X = [X^1 \, X^2] = \left[ {x}^1_1 \, {x}^1_2 \, \dots \, {x}^1_{\lvert  \mathcal{V} \rvert} \, {x}^2_1 \, {x}^2_2 \, \dots \, {x}^2_{\lvert  \mathcal{V} \rvert} \right]\) be the representations from two branches concatenated along the axis of nodes and tokens.   
Then the new representations \(\tilde{x}^b_i\)'s are computed via the multihead self-attention along the axis of concatenated nodes and tokens and residual connection:
\begin{align*}
    \hat{X} & = \text{MultiheadSelfAttention}\left( X \right) \text{ , } \\ 
    \tilde{x}^b_i & = \hat{x}^b_i + x^b_i 
    \text{ , }
\end{align*}\XL{need fix}
where the self-attention aggregates information across the newly concatenated axis of nodes and tokens. Its input \(X\) and its output \(\hat{X}\) have exactly the same dimensions, and can be indexed in the same way.

\section{Results and discussion}

\subsection{Experimental setup}\XL{random seed etc?}
We evaluate our proposed framework across five established protein benchmarks, including Model Quality Assessment (MQA), Protein-Ligand Binding Affinity (LBA), Reaction Prediction, Protein-Protein Binding Site Prediction (PPBS), and B-cell Epitope (BCE) Prediction (We defer the detail of datasets, tasks and results in \secref{sec:tasks_results}).
In these evaluations, we employ ProNet~\citep{wang2023pronet} for its \fix{GNN}
architectures and ESM-2 \citep{lin2023esm} as the pretrained protein language models ($\pLMs$). ProNet is noted for its hierarchical protein representations and an extensive message passing system that captures a complete 3D protein graph representation. A vital component of our framework involves ProNet's geometric transformation $F(\cdot)$, which guarantees SE(3) invariance. This invariance is critical for preserving the accuracy of protein representations regardless of their rotations and translations~\citep{defresne2021protein}. Ensuring this invariance is crucial for consistent and accurate comparisons of protein structures by eliminating variability due to orientation or positional differences.

Using a pretrained GNN and $\pLMs$, we conduct experiments with serial fusion~\citep{wu2023integration}, where representations from the $\pLMs$ are fed into the GNN as features. 
\fix{Both ProNet and Serial Fusion achieved previous state-of-the-art performance across various tasks and served as strong competitive baselines.}
Additionally, we explore our local \algname with gating and global \algname employing multi-head self-attention. Furthermore, for the ligand binding affinity task, we utilize the same GNN for the ligand as described in the studies we reference ~\citep{wang2023pronet, somnath2021holoprot}.

\subsection{Tasks, datasets and experiment results}\label{sec:tasks_results}

\subsubsection{Single-protein representation task}
In the single-protein representation task, which encompasses two specific tasks—Reaction Classification and Model Quality Assessment (MQA)—the results are as follows:

    \begin{table}[t!]
    \centering
    \begin{tabular}{l l c}
    \hline
    \textbf{Method}          & $\pLMs$     & \textbf{Accuracy} \\ \hline
    ProNet (Amino Acid)              & $\times$         & 0.79             \\ 
    Serial Fusion        & $\checkmark$         & 0.8105                 \\ 
    Local \algname    & $\checkmark$     & \textbf{0.8757}           \\ 
    Global \algname    & $\checkmark$    & \underline{0.8412}           \\ \hline
    \end{tabular}
    \caption{Comparison of methods on reaction prediction. GNN uses ProNet (Amino Acid) as the baseline backbone. The top two results are highlighted as \textbf{1st} and \underline{2nd}.}
    \label{tbl:react_result}
    \end{table}

\textbf{Reaction Classification.} 
 Reaction classification prediction involves determining the specific biochemical reaction catalyzed by an enzyme, a task critical for understanding metabolic pathways and designing enzyme-targeted drugs. 
Enzymes, which serve as biological catalysts, are categorized by enzyme commission (EC) numbers based on the reactions they facilitate. We utilize the dataset and experimental setup from ~\citet{hermosilla2020ieconv} to evaluate our methods. This dataset comprises 37,428 proteins across 384 EC numbers~\citep{berman2002protein,dana2019sifts}. For reaction prediction, we assess performance using accuracy metrics. 

The summarized results, presented in~\tabref{tbl:react_result}, illustrates the enhancement in performance of the ProNet-based GNN model when different fusion approaches are applied. Here's a breakdown of how each approach improves upon the standard ProNet setup: The base accuracy with ProNet alone, without integrating protein language model ($\pLMs$) features, is 0.79. This serves as the benchmark for subsequent comparisons. By incorporating $\pLMs$' features through serial fusion—where $\pLMs$ output are directly used as inputs to the GNN—accuracy improves to 0.8105. This suggests that the additional contextual information from the $\pLMs$ help refine the GNN’s predictions. Local \algname yields the highest performance, with an accuracy of 0.8757. Local \algname involves a more dynamic integration where the GNN not only uses $\pLMs$' features but also adapts how these features are combined based on the specific requirements of corresponding nodes or amino acids in different modality. This method provides a more targeted and effective use of the information from the $\pLMs$, leading to significantly improved accuracy. The accuracy of global \algname reaches 0.8412, which is the second-best result. Global \algname extends the concept by allowing information exchange across all nodes, not just corresponding ones, leveraging a multi-head self-attention mechanism. This broader scope of information exchange further enhances the model’s ability to generalize and accurately predict reactions, though it is slightly less effective than the local \algname.

Overall, the enhancements from these fusion techniques illustrate how integrating and dynamically managing additional sources of information (like those from $\pLMs$) can significantly improve the performance of a GNN in complex tasks such as reaction prediction. Both of our \algname approaches outperform the state-of-the-art methods, serial fusion, by a wide margin.

\XL{Songhao, add detail about the dataset difference of ours and pronets}

\begin{table}[t]
\centering
\footnotesize  
\setlength{\tabcolsep}{4pt}
\begin{tabular}{@{}llccc@{}}
\toprule
Method    & $\pLMs$    & MSE \(\downarrow\) & $R_p$ \(\uparrow\) & $R_s$ \(\uparrow\) \\ 
\midrule
ProNet (Amino Acid)   & $\times$ & 0.1934 \fix{$\pm$ 0.006} & 0.5479 \fix{$\pm$ 0.034} & 0.5948 \fix{$\pm$ 0.052} \\
Serial fusion & $\checkmark$  & 0.1915 \fix{$\pm$ 0.001}  & \underline{0.6024 \fix{$\pm$ 0.005}} & 0.5942 \fix{$\pm$ 0.001} \\
Local \algname & $\checkmark$  & \textbf{0.1846 \fix{$\pm$ 0.001}} & \textbf{0.6075 \fix{$\pm$ 0.006}} & \textbf{0.5969 \fix{$\pm$ 0.002}} \\
Global \algname & $\checkmark$  & \underline{0.1881 \fix{$\pm$ 0.001}}  & 0.5977 \fix{$\pm$ 0.003} & \underline{0.5961 \fix{$\pm$ 0.004}}  \\
\bottomrule
\end{tabular}
\caption{Comparison of methods on model quality assessment (MQA). GNN uses ProNet (Amino Acid) as the baseline backbone. The top two results are highlighted as \textbf{1st} and \underline{2nd}. \fix{Results on over 5 independent runs.}}
\label{tab:qa_results} 
\end{table}
    
\textbf{Model Quality Assessment.} 
Model Quality Assessment (MQA) plays a critical role in structure prediction by selecting the best structural model of a protein from a large pool of candidate structures \citep{cheng2019estimation}. For many recently solved but unreleased protein structures, structure generation algorithms produce an extensive set of candidate models. MQA methods are evaluated based on their ability to predict the Global Distance Test Total Score (GDT-TS) of a candidate structure relative to the experimentally determined structure of the target protein. The evaluation of MQA approaches often relies on databases comprising all structural models submitted to the Critical Assessment of Structure Prediction (CASP) \citep{kryshtafovych2019critical} experiments.

\SJ{Will add missing cols later. found a bug in script}
For this task, 
we evaluate mean squared error (MSE), Pearson correlation coefficient ($R_p$), and Spearman correlation coefficient ($R_s$), calculated across all decoys of all targets (R). The results are detailed in \tabref{tab:qa_results}. ProNet, at the amino acid level, serves as the GNN baseline. By integrating insights from a pretrained large language model, we observe enhanced performance on various scales, affirming that the additional contextual data from the $\pLMs$ significantly refine the GNN’s predictions. Furthermore, both of our \algname approaches surpass the serial fusion in most metrics, demonstrating that \algname, which facilitates amino acid knowledge exchange across different modalities on both local and global scales, improves upon the conventional serial fusion approach.

\begin{table*}[t]
\centering
\scalebox{1}{
\footnotesize  
\setlength{\tabcolsep}{4pt}
\begin{tabular}{@{}llcccccc@{}}
\toprule
Method & $\pLMs$  & \multicolumn{3}{c}{Sequence Identity 30\%} & \multicolumn{3}{c}{Sequence Identity 60\%} \\ \cmidrule(l){3-5} \cmidrule(l){6-8} & 
& RMSE  \(\downarrow\) & $R_p$ \(\uparrow\) & $R_s$ \(\uparrow\) & RMSE  \(\downarrow\) & $R_p$ \(\uparrow\) & $R_s$ \(\uparrow\) \\ 
\midrule
ProNet (Amino Acid) & $\times$ & 1.455 & 0.536 & 0.526 & 1.397 & 0.741 & 0.734 \\
Serial fusion& $\checkmark$  & \underline{1.402} & \underline{0.576} & \textbf{0.568} & 1.370 & 0.755 & 0.746 \\
Local \algname & $\checkmark$ & 1.404 & \textbf{0.581} & \underline{0.567} & \underline{1.323} & \underline{0.770} & \underline{0.761} \\
Global \algname & $\checkmark$ & \textbf{1.389} & 0.573 & {0.562} & \textbf{1.291} & \textbf{0.782} & \textbf{0.781} \\
\midrule
ProNet (All Atom) & $\times$ & 1.463 & 0.551 & 0.551 & 1.343 & 0.765 & 0.761 \\
Serial fusion& $\checkmark$  & 1.407 & \underline{0.600} & 0.586 & 1.332 & 0.764 & 0.760 \\
Local \algname  & $\checkmark$& \underline{1.382} & \textbf{0.611} & \underline{0.598} & \textbf{1.289} & \textbf{0.782} & \textbf{0.776} \\
Global \algname  & $\checkmark$& \textbf{1.380} & 0.598 & \textbf{0.600} & \underline{1.326} & \underline{0.775} & \underline{0.774} \\
\bottomrule
\end{tabular}}
\caption{\label{tab:mainresults} Results on protein-ligand binding affinity prediction task.  For baselines, We took the results from the paper of ProNet \citep{wang2023pronet}. The top rows all use ProNet-Amino Acid as the GNN for proteins, and the bottom rows all use ProNet-All Atom.  All our fusion models use ESM-2 for $\pLMs$. \textbf{Bolded} numbers are the best performance within the comparison group.
The top two results are highlighted as \textbf{1st} and \underline{2nd}.
}
\end{table*}

\subsubsection{Protein-molecules representation task}

\begin{figure*}[ht]
    \centering
    \includegraphics[width=1\linewidth]{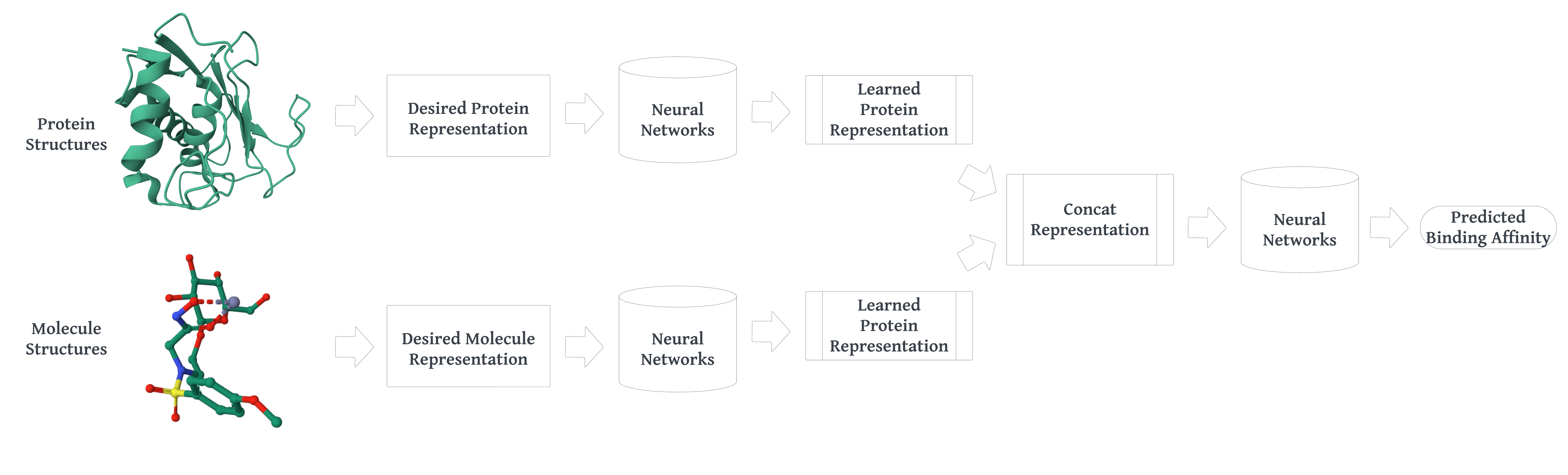}
    \caption[Interaction-free binding affinity prediction architecture with proposed fusion methods]{This figure shows the interaction-free architecture. The process begins with converting the SMILES code of a drug into a molecular graph, which is then processed by a GNN to learn a graph representation. Concurrently, the protein structure is extracted from corresponding PDB file, encoded into protein graphs, and subjected to our fusion methods for joint representation learning. The resulting representation vectors from both the drug and target protein are concatenated and fed through several fully connected layers to predict the drug–target affinity.}
    \label{fig:ba_interaction_free_specific}
\end{figure*}

\textbf{Ligand Binding affinity.} 
Protein-ligand binding affinity (LBA) prediction~\citep{liu2024binding} is a critical task in drug discovery, as it estimates the interaction strength between a candidate drug molecule and a target protein. For this study, we utilize the PDBbind database \citep{wang2005pdbbind}, a curated resource of protein-ligand complexes sourced from the Protein Data Bank (PDB), annotated with their respective binding strengths. To ensure a robust evaluation, the dataset is partitioned such that proteins in the test set share no more than 30\% or 60\% sequence identity with any protein in the training set.

The advancement of interaction-free methods as illustrated in ~\figref{fig:ba_interaction_free_specific} for binding affinity prediction emphasizes the need for sophisticated computational techniques independent of physical interaction data. These methods employ separate neural networks and representations for proteins and molecules, with proteins providing a more complex computational canvas due to their intricate structures and dynamic functions. In our study, we represent proteins using ProNet and ligands as outlined in~\citep{wang2023pronet, somnath2021holoprot}.
We evaluate these methods at both the amino-acid and all-atom levels for a comprehensive assessment of binding affinity prediction. Our evaluations involve comparing root mean squared error (RMSE), Pearson correlation ($R_p$), and Spearman correlation ($R_s$), with the main results detailed in Table~\ref{tab:mainresults}. ProNet, previously the state-of-the-art, serves as our baseline. Serial fusion improves upon this, reducing RMSE by 0.03-0.05. Our \algname approaches further decrease RMSE by 0.01-0.08, with global \algname achieving the lowest RMSE in the challenging 30\% sequence identity split, demonstrating strong generalizability.

Additionally, our methods show a 4 percentage point improvement in Pearson and Spearman correlations for the 60\% data split, and a 7 percentage point increase in the more demanding 30\% split. Across different data regimes and GNN variants, our \algname significantly surpasses previous systems and sets a new benchmark for binding affinity prediction.
Overall, the use of all-atom level information enhances performance across various metrics compared to amino acid level data. While ProNet provides a strong foundation and performs well, the integration of sequential knowledge from $\pLMs$ consistently boost performance. Our \algname approach outperforms serial fusion, underscoring its effectiveness in enhancing predictive accuracy.

\subsubsection{Protein-protein representation task}\label{sec:ppbs}

\begin{table*}[t]
\centering
\scalebox{0.82}{
\footnotesize  
\setlength{\tabcolsep}{4pt}
\begin{tabular}{@{}llcccccc@{}}
\toprule
Method & $\pLMs$   & \multicolumn{5}{c}{AUCPR \(\uparrow\)}  \\ 
\cmidrule(l){3-7}  & 
& Test (70\%) & Test (Homology) & Test (Topology) & Test (None) & Test (All)  \\ 
\midrule
ScanNet & $\times$ & 0.732  \fix{$\pm$ 0.005}  & 0.712 \fix{$\pm$ 0.004}  & \textbf{0.735 \fix{$\pm$ 0.005}} & \underline{0.605 \fix{$\pm$ 0.006}} & 0.694 \fix{$\pm$ 0.003} \\
ProNet (Amino Acid) & $\times$ & 0.817 \fix{$\pm$ 0.007} & 0.705 \fix{$\pm$ 0.004} & 0.691 \fix{$\pm$ 0.006} & 0.577 \fix{$\pm$ 0.003} & 0.685 \fix{$\pm$ 0.002} \\
Serial fusion & $\checkmark$ & 0.807 \fix{$\pm$ 0.003}  & 0.728 \fix{$\pm$ 0.002}  & 0.726 \fix{$\pm$ 0.004} & 0.592 \fix{$\pm$ 0.002} &  0.700 \fix{$\pm$ 0.003}   \\
Local \algname & $\checkmark$  & \textbf{0.839 \fix{$\pm$ 0.004}} & \textbf{0.755 \fix{$\pm$ 0.003}} & 0.714 \fix{$\pm$ 0.002} & 0.594 \fix{$\pm$ 0.003} & \textbf{0.716 \fix{$\pm$ 0.002}} \\
Global \algname & $\checkmark$  & \underline{0.828 \fix{$\pm$ 0.004}} & \underline{0.733 \fix{$\pm$ 0.002}}  & \textbf{0.735 \fix{$\pm$ 0.003}}  & \textbf{0.620 \fix{$\pm$ 0.002}} & \underline{0.714 \fix{$\pm$ 0.003}} \\
\bottomrule
\end{tabular}
}
\caption{ Performance assessment for predicting protein-protein binding sites (PPBs) is presented with the Area Under the Curve for the Precision-Recall (AUCPR) metric. The proteins in the test set are categorized into four distinct, non-overlapping groups. For the masif-site, only aggregated performance is displayed, as its training dataset differs from ours. Entries in bold highlight the top performance. The top two results are highlighted as \textbf{1st} and \underline{2nd}. \fix{The results for ScanNet are taken directly from the original paper \citep{tubiana2022scannet}, which reports performance across 10 random seeds. For our methods, we report the mean and standard deviation over 5 independent runs}. }\label{tab:ppbs_results_aucpr} 
\end{table*}

\textbf{PPBS.}
Protein-protein binding sites (PPBS) are specific protein residues crucial for high-affinity interactions (PPIs). These sites require both structural stability and specificity to the binding partner's conformation, making them challenging to predict with traditional methods due to their varied, less conserved motifs. 
PPBS is crucial for understanding disease mechanisms and designing therapeutics that target specific protein interactions. More specifically, identifying the PPBS of a protein provides valuable insights into its in vivo behavior, particularly when its interaction partners are unknown, and it can guide docking algorithms by narrowing the search space. 

For this task, we evaluate PPBS using the ScanNet metric~\citep{tubiana2022scannet} based on a detailed dataset from the Dockground database, which includes 20,000 protein chains and spans various complex types, covering 5 million amino acids, with 22.7\% identified as PPBS.
We assess performance using the area under the precision-recall curve (AUCPR). Our test set proteins, aligned with the ScanNet setup, are categorized into four exclusive groups: (a) Test 70\%: Proteins sharing at least 70\% sequence identity with training set examples. (b) Test homology: Proteins with up to 70\% sequence identity to any training example but within the same protein superfamily (H-level in CATH classification). (c) Test topology: Proteins sharing similar topology (T-level in CATH classification) but not the same superfamily. (d) Test none: Proteins that do not fit any previous categories. Our results, detailed in~\tabref{tab:ppbs_results_aucpr}, show that while ProNet was our initial benchmark and typically underperformed relative to ScanNet, employing a pretrained protein large language model via \algname markedly enhanced performance on all baseline metrics. Specifically, local \algname exceeds ScanNet in most metrics, and global \algname outperforms ScanNet in all metrics.

\begin{table}[ht]
\centering
\footnotesize  
\setlength{\tabcolsep}{4pt}
\begin{tabular}{@{}llc@{}}
\toprule
Method & $\pLMs$   & AUCPR \(\uparrow\)  \\ 
\midrule
ScanNet & $\times$ & 0.177  \\
ProNet (Amino Acid) & $\times$ & 0.1874  \\
Serial fusion & $\checkmark$ & 0.2222   \\
Local \algname & $\checkmark$  & \underline{0.2352} \\
Global \algname & $\checkmark$  & \textbf{0.2418} \\ 
\bottomrule
\end{tabular}
\caption{Performance assessment for B-cell conformational epitopes (BCE). 
is presented with the Area Under the Curve for the Precision-Recall (AUCPR) metric. The proteins in the test set are categorized into four distinct, non-overlapping groups. For the masif-site, only aggregated performance is displayed, as its training dataset differs from ours. Entries in bold highlight the top performance. The top two results are highlighted as \textbf{1st} and \underline{2nd}.}
\label{tab:antibody_results_aucpr} 
\end{table}

\textbf{Prediction of BCEs (B cell epitopes).}\XL{revisit}
B-cell conformational epitopes (BCEs) are residues that are actively involved in the interaction between an antibody and an antigen. Prediction of BCEs (B cell epitopes), also known as discontinuous epitopes, are regions on antigens recognized by B-cell receptors (BCRs) or antibodies where the amino acids that make up the epitope are not contiguous along the primary sequence but come together in the three-dimensional space due to the folding of the protein. While theoretically, any surface residue could trigger an immune response, certain residues are more favorable because antibodies targeting these residues can be more easily matured to achieve high specificity and affinity. 

This in contrast to protein-antibody binding sites prediction that protein-antibody binding sites can involve both conformational and linear epitopes. 
Exhaustive and high-throughput experimental identification of BCEs is difficult due to their distribution over various noncontiguous segments of protein. 
Predicting BCEs presents challenges due to their evolutionary instability and the absence of comprehensive epitope maps for specific antigens. Nevertheless, in silico prediction of BCEs can be effectively used to develop epitope-based vaccines and to create therapeutic proteins that do not trigger an immune response.

In this study, we adopted the dataset configuration from Scannet~\citep{tubiana2022scannet}, and extracted data from the SabDab database~\citep{dunbar2014sabdab}. This dataset includes 3,756 protein chains, each annotated with BCEs, where 8.9\% of the residues are identified as BCEs—a figure that likely underestimates the actual percentage. The dataset was segmented into five subsets for cross-validation, with each pair of sequences from different subsets having no more than 70\% sequence identity.
\tabref{tab:antibody_results_aucpr}  illustrates the performance evaluation of BCE prediction using the Area Under the Precision-Recall Curve (AUCPR). The results indicate that both the local and global \algname approaches surpass the serial fusion method in effectiveness. Notably, the Global \algname approach achieved a score of 0.2418, significantly outperforming all other baselines, including the current state-of-the-art, ScanNet, by a considerable margin.

\section{Limitations.}
We believe the framework is versatile and adaptable to various future GNN and $\pLMs$, and can benefit other GNNs and $\pLMs$ for other downstream tasks involving proteins.  One constraint of our framework is that it requires that the $\pLMs$ and GNN somehow represent nodes of the graph at the same level of, and does not yet have a way to utilize structures of systems with multi-scale representations.  We leave this extension to future work.

\section{Conclusion.}
\label{sec:conclusion}
In this work, we introduce the \algname framework, a novel fusion architecture for protein representation learning that harnesses the complementary strengths of protein language models ($\pLMs$) and graph neural networks (GNNs). By integrating both sequential and structural perspectives, \algname enhances protein representations for a variety of prediction tasks.
Building on this framework, we propose two variants: local-\algname, which incorporates a gating mechanism to enable bidirectional and hierarchical information exchange between related nodes (e.g., backbone and amino acid); and global-\algname, which employs multi-head attention to facilitate broader bidirectional and hierarchical interactions across diverse nodes.
To evaluate the effectiveness of \algname, we conduct experiments across three tasks: single-protein representation, protein–molecule interaction, and protein–protein interaction. Our approaches consistently outperform state-of-the-art methods, including Pronet and the Serial Fusion approach. These results demonstrate that both variants of \algname significantly advance performance, highlighting the advantages of merging representations from the $\pLMs$ and GNN perspectives in a bidirectional and hierarchical manner.

\subsubsection*{Acknowledgements}

This work is supported in part by the RadBio-AI project (DE-AC02-06CH11357), U.S. Department of Energy Office of Science, Office of Biological and Environment Research, the Improve project under contract (75N91019F00134, 75N91019D00024, 89233218CNA000001, DE-AC02-06-CH11357, DE-AC52-07NA27344, DE-AC05-00OR22725), 
the Exascale Computing Project (17-SC-20-SC), a collaborative effort of the U.S. Department of Energy Office of Science and the National Nuclear Security Administration.

\bibliography{reference}

\clearpage

\onecolumn
\appendix
\section{Appendix}
\subsection{Computing infrastructure and wall-time} \label{app:computing_infrastructure}
We conducted experiments on a platform where we can access CPU nodes (approximately 120 cores) and GPU nodes (approximately 10 Nvidia V100 GPUs). 
Using a single Nvidia V100, our wall-time is about 24 hours to train one single model.

\subsection{Hyperparameters and architectures} \label{app:hyperparameters}
Table \ref{app:tab:hyperparams} provides a list of hyperparameter ranges we used or searched among for our experiments.

\begin{table}[ht]
\centering
\scriptsize
\begin{tabular}{@{}ll@{}}
\toprule
Hyperparameter       & Value or range \\ \midrule
\emph{General}             &                \\
Learning rate        &     $[1e-5, 1e-4]$           \\
cutoff             &       $[4,6,8,10]$          \\
Dropout            &       $[0.2, 0.3, 0.5]$         \\
Batch Size            &       $[16, 32]$         \\
ESM Size (\# of Layers)           & $[6,12,30]$ \\
\# of Epochs training &        256        \\
Gaussian noise &        [True, False]        \\
Euler noise  &        [True, False]        \\
                     &                \\
\emph{Serial}                &                \\
Hidden dimension     &    $[125,256]$            \\
\# of layers &   $[3,4,5]$             \\
\bottomrule
\end{tabular}
\caption{\label{app:tab:hyperparams} {{Hyperparameters}}. }
\end{table}

\begin{itemize}
    \item Gaussian noise: if True, will add noise to the node features before each interaction block \citep{wang2023pronet}.
    \item Euler noise: if True, will add noise to Euler angles \citep{wang2023pronet}
\end{itemize}

\fix{\subsection{Ablation Study: Number of ESM layers}} \label{app:ablation:esm_layer}

\fix{
\begin{table}[t]
\centering
\footnotesize  
\setlength{\tabcolsep}{4pt}
\begin{tabular}{@{}llccc@{}}
\toprule
ESM Layer    & \# of Parms    & MSE \(\downarrow\) & $R_p$ \(\uparrow\) & $R_s$ \(\uparrow\) \\ 
\midrule
6   & 8M & 1.377 & 0.736 & 0.741 \\
12 & 35M  & 1.365  & 0.739 & 0.736	 \\
30  & 150M  & 1.323 & 0.770 & 0.761 \\
\bottomrule
\end{tabular}
\caption{\fix{Binding Affinity 60\% with Local Bi-Hierarchical Fusion. Ablation study showing that increasing ESM layers improves performance}}
\label{tab:ablation:esm_layer} 
\end{table}
}

\subsection{Dataset Description} \label{app:dataset}
We describe all datasets used in the main text here. Notably, since ESM2 was trained on sequences cropped to a maximum length of 1024, the model—particularly its learned positional embeddings—is unable to process longer sequences. Therefore, we exclude all protein sequences exceeding 1024 residues from our dataset. Additionally, proteins for which PDB files are not available are excluded.

\subsubsection{Reaction classification}
Enzymes, which act as biological catalysts, are classified by Enzyme Commission (EC) numbers according to the reactions they catalyze. To evaluate our approach, we adopt the dataset and experimental protocol introduced by \citet{hermosilla2020ieconv}. The dataset is divided into 25670 proteins for training, 2852 for validation, and 5598 for testing. Each EC number is present across all three splits, and protein chains sharing more than 50\% sequence similarity are grouped together.

\subsubsection{Model quality assessment}

The Critical Assessment of Structure Prediction (CASP) \citep{cheng2019estimation} is a long-running international competition focused on protein structure prediction, with CASP14 being the latest edition. In this challenge, newly resolved experimental structures are withheld to evaluate predictive performance. Following the protocol of \citep{wu2023integration}, we divide the decoy sets by target and release year: CASP11-12 are used for training, CASP13 for validation, and CASP14 for testing. This results in a train/validation/test split of 22885/3536/6492.

\subsubsection{Ligand binding affinity}
PDBbind includes X-ray crystal structures of proteins complexed with small molecules and peptide ligands. We utilize a dataset derived from PDBbind \citep{wang2005pdbbind}, which is available in two variants based on sequence identity thresholds of 30\% and 60\%. The 30\% identity split yields train/validation/test sets of 2929/325/480, while the 60\% identity split produces train/validation/test sets of 3016/335/424.

\subsubsection{Protein-protein binding sites}
For the tasks of protein–protein binding site (PPBS) prediction and B-cell conformational epitope (BCE) prediction, we utilize the dataset provided by ScanNet \citep{tubiana2022scannet}. As described in Section~\ref{sec:ppbs}, ScanNet defines four distinct test groups for PPBS evaluation. Instead of using separate validation sets for each group as defined in ScanNet, we derive a single validation set from the training partition. This results in 11225 proteins for training, 1247 for validation, and the following counts for testing: 539 for Test 70\%, 1453 for Test Homology, 893 for Test Topology, 1050 for Test None, and 3935 for Test All.

For the BCE task, the dataset is divided into five subsets for cross-validation, with the constraint that no two subsets share more than 70\% sequence identity. In our experiments, we use the first three subsets for training, and the remaining two for validation and testing, resulting in a train/validation/test split of 2106/914/485.

\end{document}